%% file: cvpr.tex
\documentclass[10pt,twocolumn,letterpaper]{article}

\usepackage{cvpr}              
\usepackage{float}


\definecolor{cvprblue}{rgb}{0.21,0.49,0.74}
\usepackage[pagebackref,breaklinks,colorlinks,allcolors=cvprblue]{hyperref}

\def\paperID{13637} 
\def\confName{CVPR}
\def\confYear{2025}

\title{SAFER: \textbf{S}harpness \textbf{A}ware layer-selective \textbf{F}inetuning for \textbf{E}nhanced \textbf{R}obustness in vision transformers}

\author{Bhavna Gopal$^1$, Huanrui Yang$^2$, Mark Horton$^1$, Yiran Chen$^1$\\ $^1$Department of Electrical and Computer Engineering, Duke University \\
$^2$Department of Electrical Engineering and Computer Science, University of California, Berkeley\\
$^1$\{bhavna.gopal, mark.horton, yiran.chen\}@duke.edu, $^2$huanrui@berkeley.edu}

\begin{document}
\maketitle
\input{0_abstract}    
\input{intro}

\input{related}

\input{method}
\input{results}
\newpage
\clearpage
{
    \small
    \bibliographystyle{ieeenat_fullname}
    \bibliography{cvpr}
}



\input{appendix}  

\end{document}

%% file: 0_abstract.tex
\begin{abstract}
Vision transformers (ViTs) have become essential backbones in advanced computer vision applications and multi-modal foundation models. Despite their strengths, ViTs remain vulnerable to adversarial perturbations, comparable to or even exceeding the vulnerability of convolutional neural networks (CNNs). Furthermore, the large parameter count and complex architecture of ViTs make them particularly prone to adversarial overfitting, often compromising both clean and adversarial accuracy.
This paper mitigates adversarial overfitting in ViTs through a novel, layer-selective fine-tuning approach: \textbf{SAFER}. Instead of optimizing the entire model, we identify and selectively fine-tune a small subset of layers most susceptible to overfitting, applying sharpness-aware minimization to these layers while freezing the rest of the model. Our method consistently enhances both clean and adversarial accuracy over baseline approaches. Typical improvements are around 5\%, with some cases achieving gains as high as 20\% across various ViT architectures and datasets.



\end{abstract}

%% file: intro.tex
\section{Introduction}
\label{sec:intro}
Vision Transformers (ViTs) \cite{dosovitskiy2020image, liu2021swin, touvron2021training} have significantly advanced computer vision architecture design, achieving state-of-the-art performance across diverse tasks, including semantic segmentation \cite{li2022bevformer}, object detection \cite{carion2020end}, and image generation \cite{peebles2023scalable}. However, as ViTs see increased deployment in real-world applications, concerns regarding their robustness against adversarial attacks—small, carefully crafted modifications to input images that mislead the model \cite{goodfellow2015explaining, szegedy2014intriguing}—have become paramount. Initial optimism that ViTs might inherently offer greater robustness than convolutional neural networks (CNNs) was quickly tempered by stronger threat models, which revealed their vulnerability to adversarial attacks at levels comparable to, or even exceeding, those of CNNs \cite{bai2021transformersrobustcnns}.
To defend against adversarial examples, ViTs continue to rely heavily on adversarial training (AT) \cite{athalye2018obfuscated, madry2019deep, croce2020reliable}, which incorporates adversarial examples into the training process to improve robustness. However, the large parameter counts and intricate architectures of ViTs exacerbate overfitting during adversarial training, limiting improvements in both adversarial robustness and clean data performance.

Recent methods, such as Attention Random Dropping (ARD) and Perturbation Random Masking (PRM) \cite{mo2022adversarialtrainingmeetsvision}, seek to address these limitations and improve AT for transformers, though they often yield inconsistent results across various settings. A more systematic approach rooted in sharpness-aware minimization (SAM) \cite{sam} provides a theoretically grounded optimizer to mitigate overfitting during training. Although SAM has shown benefits for enhancing robustness without compromising clean performance (in both CNNs and ViTs \cite{sam_and_trans}), work in this area remains relatively underexplored. Furthermore, integrating SAM into ViTs’ complex adversarial training processes can still hinder convergence, ultimately diminishing performance and failing to address overfitting effectively.

Research advances in CNNs suggest that harnessing specific architectural properties within models can significantly improve robustness. Approaches such as RiFT~\cite{rift}, CLAT~\cite{clat}, and AutoLoRA~\cite{autolora} leverage metrics and heuristics to identify and selectively exploit architecture-specific features that are critical to model robustness. For instance, CLAT utilizes hidden feature-based robust criticality indices to pinpoint "critical" layers—those disproportionately contributing to adversarial vulnerability—and fine-tunes these layers to enhance robustness.
Motivated by this insight and the challenges of performing adversarial training on the full ViT model, we hypothesize and demonstrate that ViTs, like CNNs, contain layers critical to learning the adversarial training objective effectively.
Accordingly, this work aims to identify a select subset of ViT layers that can train and converge more effectively and smoothly than the full model, yet still contribute substantially to the model's overall robust generalization. By pinpointing and selectively fine-tuning these layers, we enable targeted adjustments that maximize robustness and improve model performance on both adversarial and clean data.

Unfortunately, transformers present unique challenges for critical layer identification, making CNN-based methods like RiFT and CLAT less effective, if not completely ineffective. First, transformers' higher parameter counts exacerbate overfitting, necessitating more explicit regularization. Moreover, ViTs consist of diverse layer types—such as attentions, projections, and MLPs—that produce output distributions not directly comparable across layers, complicating the task of accurately identifying which layers are the most critical \cite{vaswani2023attentionneed}. These challenges underscore the need for a transformer-specific approach capable of effectively harnessing critical layers and enhancing adversarial robustness.

\begin{figure}[tb]
\includegraphics[scale=0.3]{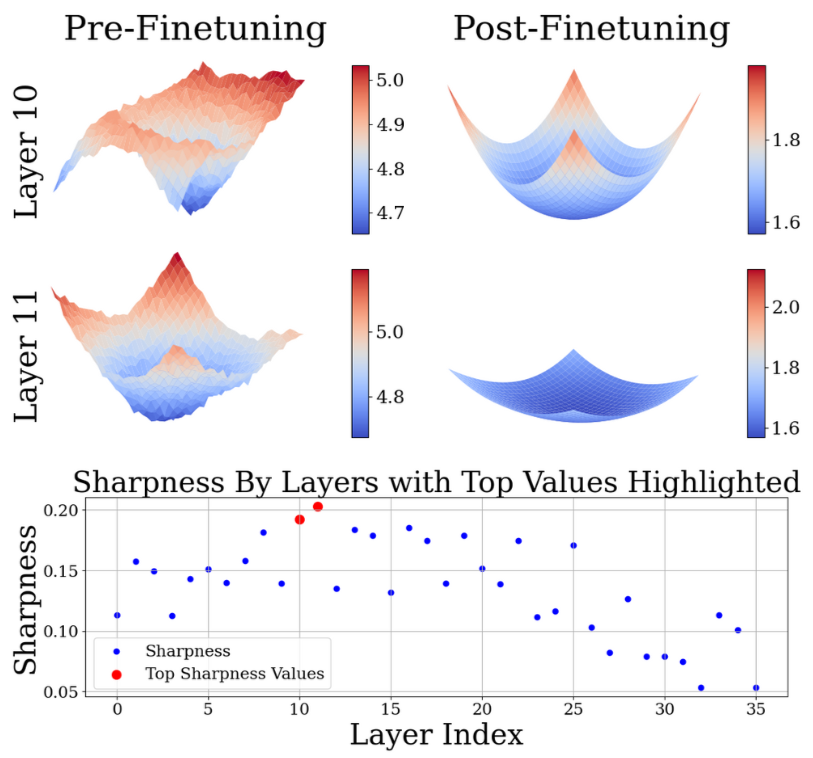}
\caption{Sharpness value (Equ.~(\ref{equ:layersharpsim})) measured over layers of an adversarially-trained DeiT-Tiny model (bottom), where the layers with top-2 values are selected for SAFER finetuning. The layers' adversarial loss landscape before adversarial finetuning (left) and after SAFER finetuning (right) are visualized on the top.}
\label{fig:surfaces}
\end{figure}

To overcome these limitations and mitigate overfitting concerns, we introduce a novel metric for critical layer identification in transformers. As shown in Figure~\ref{fig:surfaces}, we leverage insights from SAM~\cite{sam} and introduce a sharpness-based metric that precisely identifies layers most prone to overfitting, enabling targeted regularization. Building on this foundation, we propose SAFER, an adaptation of the SAM framework that mitigates overfitting specifically within these critical layers while freezing the rest of the model, ultimately enhancing both adversarial robustness and clean accuracy.

Additionally, given the widespread adoption of Parameter Efficient Fine-Tuning (PEFT) techniques for transformers \cite{peftlargemodels}, we extend SAFER to PEFT methods such as LoRA \cite{hu2021loralowrankadaptationlarge} and DORA \cite{dora}. By incorporating our algorithm across these frameworks, we demonstrate that SAFER tuning consistently enhances the robustness of transformer models, making them well-suited for diverse applications. This integration underscores the adaptability and broad utility of our approach within the transformer landscape.

Our contributions are summarized as follows: 
\begin{itemize} 
\item We introduce a novel metric for transformers that accurately identifies layers prone to overfitting in the adversarial training process.
\item We present SAFER, an advanced SAM-based fine-tuning algorithm designed to mitigate overfitting specifically within critical transformer layers, enhancing both adversarial robustness and clean accuracy. 
\item We demonstrate SAFER's versatility through results across different training methods and its integration with PEFT frameworks, highlighting its robust performance across varied scenarios. 
\end{itemize}

SAFER demonstrates consistent improvements in both adversarial and clean performance, with typical gains around 5\% and peaks of up to 20\%, achieving state-of-the-art robustness across a variety of models and datasets.

%% file: related.tex
\section{Related Work}

\subsection{Adversarial training and overfitting}
The concept of Adversarial Training (AT) emerged in response to the vulnerability of deep learning models to adversarial examples, first highlighted by \citet{goodfellow2015explaining} in CNNs. By incorporating adversarial examples into the training process, \citet{goodfellow2015explaining} demonstrated a significant improvement in model robustness.  This idea was further developed into Projected Gradient Descent Adversarial Training (PGD-AT) \citep{madry2019deep}, a minimax optimization approach that leverages multi-step PGD attacks on training data to enhance empirical robustness \citep{carlini2017evaluating, athalye2018obfuscated, croce2020reliable}. Later advancements, such as TRADES \citep{trades}, introduced a loss function to balance accuracy and robustness, while other works further bolstered resilience using model ensembling and data augmentation \citep{xie2020adversarial, yang2020dverge, carmon2022unlabeled}. Despite these improvements, techniques like PGD-AT and TRADES remain computationally demanding and prone to overfitting \citep{shafahi2019adversarial}.

For transformers, adversarial training introduces even greater challenges. Transformers’ complex architectures and high parameter counts make them especially prone to overfitting, leading to difficulties with stability and convergence. \citet{recipe_trans} attempt to address these challenges by providing tailored guidelines and recipes to improve adversarial training for transformers, though the issue of overfitting remains persistent. The recent incorporation of Sharpness-Aware Minimization (SAM) into adversarial training has shown promise, with \citet{sam_and_trans} demonstrating improved performance by smoothing the optimization landscape. However, even with SAM, transformers continue to struggle with catastrophic overfitting and convergence issues. Furthermore, these methods operate at a global level across the entire architecture, without leveraging transformer-specific architectural components.

\subsection{Layer-selective adversarial training}
Overfitting remains a central challenge in adversarial training. 
Subsequent efforts introduced methods specifically aimed at mitigating catastrophic overfitting, though these approaches were largely developed within the CNN space. RiFT \citep{rift} improved general performance by exploiting layer redundancies, though its reliance on heuristic-based redundancy measurements limited its adaptability. \citet{autolora} tackled overfitting by disentangling natural and adversarial objectives, yet model-wide adjustments constrained overall robustness. A more targeted solution, CLAT \cite{clat}, employed theoretically grounded critical layer selection to fine-tune only a subset of layers, achieving notable gains in clean and adversarial robustness. Unlike global approaches, CLAT’s dynamic selection mechanism outperformed prior methods while remaining attack-agnostic, further suggesting that layer criticality could play an essential role in enhancing robustness. However, as our empirical results demonstrate (see Table~\ref{tab:ft_comp}), the criticality index selection in CLAT fails to capture essential layers in transformer architectures due to diverse layer types.

To overcome these limitations, we introduce SAFER—a novel SAM-based approach that uniquely identifies and selectively fine-tunes only the transformer layers most prone to overfitting. By targeting these critical layers, SAFER effectively addresses overfitting and delivers robust generalization, meeting the specific challenges of adversarial training in transformer architectures.

%% file: method.tex
\section{Methods}
\label{method}

To address overfitting challenges in adversarial training for vision transformers, we first identify layers most susceptible to overfitting during training and then apply targeted layer-wise fine-tuning to reduce this tendency. In this section, we start by deriving metrics to identify overfitting layers, followed by a discussion of the training objective designed to mitigate overfitting in these layers.

\subsection{Adversarial overfitting measurement}
\label{metric}

Given a neural network model, the extent of overfitting is determined by the weights it converges to, measurable through the sharpness of the loss landscape in the locality of the converged model. Previous work, Sharpness-Aware Minimization (SAM)~\cite{sam}, derived a theorem stating that the generalization gap between the test loss \( L_{D} \) on a distribution \( \mathcal{D} \) and the model training loss \( L_{S} \) on a specific training set \( \mathcal{S} \) generated from \( \mathcal{D} \) is upper-bounded with high probability, for any \( \rho > 0 \), by

\begin{equation}
    L_{D}(w)-L_{S}(w)\leq \max_{||\epsilon||_2\leq\rho}L_{S}(w+\epsilon) - L_{S}(w) +h,
\end{equation}
where $w$ is the model weight and $h$ is a strictly increasing function of $\frac{||w||_2^2}{\rho^2}$. The theory leads to an effective sharpness measurement
\begin{equation}
\label{equ:sharp}
    \gamma = \max_{||\epsilon||_2\leq\rho}L_{S}(w+\epsilon) - L_{S}(w),
\end{equation}
where a larger sharpness $\gamma$ indicates greater susceptibility to overfitting, while a smaller $\gamma$ reduces the generalization gap, as shown in SAM~\cite{sam}.

However, given the complexity of vision transformers and the optimization difficulty of the minimax adversarial training objective, directly applying SAM to the full model during adversarial training could impede effective convergence. To this end, we propose shifting away from training all layers together. Previous research not only identifies that a small subset of CNN layers are more prone to learning non-robust features~\cite{clat,rift}, but also provides insights into how to locate these layers. Inspired by this, we propose measuring each layer's contribution to model overfitting to identify those layers that are most susceptible.

To establish a unified overfitting measure across layers of varying types and sizes, we isolate each layer's contribution to overfitting by freezing the rest of the model. Specifically, we analyze the curvature of the loss landscape when fine-tuning each individual layer with adversarial training. For an adversarial training loss $L_{adv}(w,x)$ defined as
\begin{equation}
\label{equ:adv}
    L_{adv}(w,x) = \sup_{||\delta||_p\leq d} L(w, x+\delta),
\end{equation}
where $L$ is the clean training loss, $x$ is the training data, $d$ is the attack strength, and $\delta$ is an adversarial perturbation achieved with some attack algorithm. We define the sharpness of the weight $w_i$ in layer $i$ based on Equ.~(\ref{equ:sharp}) as

\begin{equation}
\label{equ:layersharp}
    \gamma_i = \sum_{x\in\mathcal{B}} \max_{||\epsilon||_2\leq\rho}L_{adv}(w_i+\epsilon,x) - L_{adv}(w_i,x),
\end{equation}
where $\mathcal{B}$ is a batch of training data.

To reduce the cost of explicitly solving the maximization problem in Equ.~(\ref{equ:layersharp}), we can further simplify the layer sharpness formulation by applying first-order Taylor expansion as
\begin{equation}
    L_{adv}(w_i+\epsilon,x) \approx L_{adv}(w_i,x) + \epsilon^T \frac{\partial L_{adv}(w_i,x)}{\partial w_i},
\end{equation}
which yields a simplified sharpness measure as
\begin{equation}
\label{equ:layersharpsim}
    \gamma_i \approx \sum_{x\in\mathcal{B}} \max_{||\epsilon||_2\leq\rho}\epsilon^T \frac{\partial L_{adv}(w_i,x)}{\partial w_i}\propto\sum_{x\in\mathcal{B}}||\frac{\partial L_{adv}(w_i,x)}{\partial w_i}||_2.
\end{equation}

We compute sharpness measures $\gamma_i$ for all layers in the model using a single backward pass with the adversarial training loss as in Equ.~(\ref{equ:adv}). Note that all layers can be measured with the same batch of adversarial examples, requiring minimal computational overhead compared to the adversarial training process.

We rank each layer's sharpness and select the top-K layers—those most prone to overfitting—for further fine-tuning, as described in the following section.

\subsection{Layer-specific finetuning}
To address adversarial overfitting in the selected top-K layers, we apply sharpness-aware minimization (SAM) to these layers only, keeping the remaining layers frozen. Freezing all other layers reduces optimization complexity and facilitates smoother model convergence. Based on the formulation in~\cite{sam}, we convert the adversarial training objective in Equ.~(\ref{equ:adv}) into
\begin{equation}
\label{equ:sam}
    \min_{w_i} \sum_{x\in \mathcal{S}} \sup_{||\delta||_p\leq d} L(w_i+\epsilon, x+\delta),
\end{equation}
where $\epsilon = \rho \nabla_{w_i}L_{adv}(w,x)/||\nabla_{w_i}L_{adv}(w,x)||_2$ and $\mathcal{S}$ is the training set. In adversarial training, note that perturbing the weight by $\epsilon$ may result in the optimal adversarial example for weight $w_i+\epsilon$ differing from that for weight $w_i$. To save computation time (from computing adversarial samples twice) and because the weight perturbation $\epsilon$ is small, we use the adversarial sample computed on the unperturbed weight to compute the SAM loss in  Equ.~(\ref{equ:sam}). Compared to standard adversarial training, this approach requires one additional backpropagation per optimization step to compute the weight perturbation $\epsilon$ for SAM. However, this added overhead is acceptable given the multi-step optimization required to optimize the adversarial example during each adversarial training step.

Overall, the training method for SAFER is designed to systematically address overfitting in adversarial training by selectively focusing on layers most prone to it. Initially, we conduct standard adversarial training for some epochs, allowing all layers to learn meaningful features and progress toward a stable minimum. We then initiate an iterative process: identifying the top-K sharpest layers, fine-tuning these selected layers with the SAFER objective while freezing the remaining layers for several epochs, and periodically repeating sharpness measurements to update our selection of layers most susceptible to overfitting. This approach continuously targets the fine-tuning effort toward those layers most vulnerable to overfitting. The number of epochs for the initial adversarial training, along with the interval between each round of sharpness measurement and layer selection, are SAFER hyperparameters, which we explore in ablation studies in \cref{sec:ablation}.


%% file: results.tex
\section{Experiments}
\subsection{Experimental Settings}
\paragraph{Datasets and models}

We conducted experiments on three widely recognized image classification datasets: CIFAR-10, CIFAR-100, and Imagenette. CIFAR-10 and CIFAR-100 consist of 60,000 color images at a resolution of 32×32 pixels, divided into 10 and 100 classes, respectively. For Imagenette, a 10-class subset of ImageNet-1K, we chose version v1 over the latest (v2) to avoid potential data leakage caused by class reshuffling in v2. Imagenette-v1 preserves a clear separation between training and validation sets, ensuring a more reliable evaluation \citep{data, imagenette, recipe_trans}.

For our experiments, we deployed a suite of network architectures across varying sizes, including ViT~\cite{dosovitskiy2020image}, DeiT~\cite{touvron2021training}, ConViT~\cite{d2021convit}, and Swin Transformers~\cite{liu2021swin}. We closely follow the settings outlined in \citet{recipe_trans} aligning with best practices for evaluating adversarial robustness in transformers. Additionally, to ensure reliable robustness measurements, each experiment was conducted at least 10 times, with the lowest observed accuracies reported.

\paragraph{Training and Evaluation}
For all experiments, except those using exclusively standard PGD-AT or SAFER throughout training, models were first pretrained on clean data for fewer than 10 epochs, followed by adversarial training with PGD-AT for 50 epochs. In experiments where either standard PGD-AT or SAFER was applied across all epochs, models were trained without any clean data pretraining. Since SAFER can be layered over different adversarial training methods, results incorporating SAFER are denoted in our tables as "X + SAFER," where "X" refers to the baseline method applied prior to SAFER finetuning.

During adversarial training, we generated adversarial examples using PGD with a random start \citep{madry2019deep}, setting an attack budget of \( \epsilon = 0.03 \) under the \( \ell_\infty \) norm, a step size of \( \alpha = 0.007 \), and using 20 attack steps. For PGD-AT, we employed either the SAM or SGD optimizer during training, with the optimizer indicated in brackets, such as PGD-AT (SGD). These same settings were maintained for PGD-based attack evaluations.

To further assess robustness, we conducted evaluations using AutoAttack, a comprehensive ensemble-based method that combines multiple attack types, including two PGD variants, the FAB attack, and Square Attack \cite{croce2020reliable}.

Unless otherwise noted, these training and evaluation settings remained consistent across all experiments.

All experiments were conducted on NVIDIA RTX A5000 GPUs. Fine-tuning began with an initial learning rate of 0.015, following the decay schedule from \citet{sam}, but with a modified decay factor, reduced from 5 to 2. Standard data augmentations, including random cropping with padding and random horizontal flipping, were applied.

\paragraph{SAFER settings}
Equ.~(\ref{equ:layersharpsim}) shows the computation for layer sharpness that guides our layer selection process. In customizing the SAFER methodology for different network sizes, we designate approximately 5\% of layers as critical, based on hyperparameter optimization.
To refine our approach adaptively, we dynamically adjust the selected layers for fine-tuning by recalculating sharpness every ten epochs.

\subsection{Comparative Performance}

\begin{table*}[htbp]
\caption{Consolidated Performance Comparison of Various Models on CIFAR-10, CIFAR-100, and Imagenette with Clean, PGD-20, and Auto Attack (AA) Accuracy.}

\label{tab:wb}
\centering
\footnotesize
\begin{tabular}{l@{\hspace{5pt}}l@{\hspace{10pt}}c@{\hspace{15pt}}c@{\hspace{15pt}}c@{\hspace{10pt}}c@{\hspace{15pt}}c@{\hspace{15pt}}c@{\hspace{10pt}}c@{\hspace{15pt}}c@{\hspace{15pt}}c}
\toprule
\textbf{Model} & {Method} & 
\multicolumn{3}{c}{\textsc{Cifar-10}} & \multicolumn{3}{c}{\textsc{Cifar-100}} & \multicolumn{3}{c}{\textsc{Imagenette}} \\
\cmidrule(lr){3-5} \cmidrule(lr){6-8} \cmidrule(lr){9-11}
 &  & \textsc{Clean} & \textsc{PGD-20} & \textsc{AA} & \textsc{Clean} & \textsc{PGD-20} & \textsc{AA} & \textsc{Clean} & \textsc{PGD-20} & \textsc{AA} \\
\midrule

\textbf{DeiT-Ti}~\cite{touvron2021training} & \textsc{pgd-at (sgd)} & 75.46 & 48.10 & 43.62 & 53.11 & 27.97 & 25.45 & 80.60 & 56.00 & 53.80 \\
 & \textsc{pgd-at (sam)} & 77.12 & 54.45 & 45.12 & 56.88 & 32.83 & 32.00 & 83.91 & 65.42 & 65.20 \\
 & \textsc{ard + prm} & 79.60 & 50.33 & 45.99 & 54.67 & 30.67 & 30.02 & 90.40 & 65.00 & 64.00 \\
  & \textsc{\citet{tian2022deeper}} & 75.50 & 46.33 & 42.10 & 52.67 & 27.45 & 23.33 & 82.88 & 57.24 & 54.89 \\
 & \textsc{\cite{tian2022deeper} + safer} & 77.21 & 48.05 & 44.67 & 54.04 & 29.31 & 28.99 & 83.79 & 50.80 & 56.38 \\
 & \textbf{\textsc{safer}} & \textbf{82.36} & \textbf{68.50} & \textbf{50.12} & \textbf{62.37} & \textbf{40.15} & \textbf{35.65} & \textbf{92.45} & \textbf{68.36} & \textbf{67.88} \\
\midrule
\textbf{DeiT-S} & \textsc{pgd-at (sgd)} & 81.43 & 51.88 & 47.10 & 55.36 & 29.12 & 27.88 & 92.20 & 64.60 & 63.40 \\
 & \textsc{pgd-at (sam)} & 82.10 & 53.64 & 46.55 & 58.81 & 35.44 & 31.03 & 92.00 & 67.09 & 65.12 \\
 & \textsc{ard + prm} & 83.04 & 52.52 & 48.34 & 58.45 & 30.13 & 28.15 & 91.00 & 66.60 & 65.80 \\
 & \textbf{\textsc{safer}} & \textbf{86.01} & \textbf{70.26} & \textbf{51.58} & \textbf{63.66} & \textbf{42.29} & \textbf{36.70} & \textbf{94.78} & \textbf{69.86} & \textbf{67.20} \\
\midrule
\textbf{ViT-S}~\cite{dosovitskiy2020image} & \textsc{pgd-at (sgd)} & 79.59 & 50.86 & 46.37 & 55.01 & 27.45 & 23.21 & 90.40 & 63.80 & 62.80 \\
 & \textsc{pgd-at (sam)} & 80.11 & 52.10 & 48.11 & 56.45 & 29.30  & 25.52 & 90.00 & 65.11 & 62.13 \\
 & \textsc{ard + prm} & 81.86 & 51.73 & 47.33 & 58.55 & 30.21 & 24.46 & 91.40 & 65.20 & 63.00 \\
 & \textbf{\textsc{safer}} & \textbf{83.40} & \textbf{68.89} & \textbf{50.12} & \textbf{60.24} & \textbf{32.56} & \textbf{25.50} & \textbf{94.22} & \textbf{67.01} & \textbf{64.57} \\
\midrule
\textbf{ViT-B} & \textsc{pgd-at} & 83.16 & 52.98 & 49.06 & 55.22 & 29.31 & 24.45 & 93.40 & 68.80 & 67.00 \\
 & \textsc{pgd-at (sam)} & 84.45 & 53.63 & 51.84 & 57.18 & 30.07 & 26.42 & 94.58 & 69.47 & 68.12 \\
 & \textsc{ard + prm} & 84.90 & 53.80 & 50.03 & 59.80 & 31.24 & 27.12 & 95.00 & 70.00 & 69.60 \\
 & \textbf{\textsc{safer}} & \textbf{86.12} & \textbf{71.95} & \textbf{53.51} & \textbf{61.55} & \textbf{33.19} & \textbf{29.89} & \textbf{97.65} & \textbf{72.43} & \textbf{70.34} \\
\midrule

\textbf{ConViT-Ti}~\cite{d2021convit} & \textsc{pgd-at (sgd)} & 53.09 & 33.63 & 29.65 & 40.45 & 20.22 & 19.83 & 63.60 & 39.20 & 36.60 \\
 & \textsc{ard + prm} & 80.28 & 47.47 & 45.42 & 55.64 & 26.67 & 26.60 & 90.40 & 65.00 & 64.40 \\
 & \textbf{\textsc{safer}} & \textbf{83.45} & \textbf{51.72} & \textbf{48.19} & \textbf{59.10} & \textbf{28.34} & \textbf{28.20} & \textbf{92.28} & \textbf{68.07} & \textbf{66.97} \\
\midrule
\textbf{ConViT-S} & \textsc{pgd-at (sgd)} & 54.03 & 34.61 & 30.60 & 44.75 & 20.23 & 22.12 & 87.40 & 64.20 & 61.60 \\
 & \textsc{pgd-at (sam)} & 55.92 & 38.65 & 45.02 & 46.12 & 21.10 & 23.04 & 88.05 & 65.03 & 65.58 \\
 & \textsc{ard + prm} & 84.32 & 53.10 & 48.85 & 58.32 & 25.33 & 27.99 & 94.40 & 68.20 & 67.60 \\
 & \textbf{\textsc{safer}} & \textbf{87.34} & \textbf{56.55} & \textbf{50.00} & \textbf{61.01} & \textbf{30.45} & \textbf{29.06} & \textbf{96.21} & \textbf{71.43} & \textbf{69.31} \\
\midrule

\textbf{ConViT-B} & \textsc{pgd-at (sgd)} & 61.54 & 38.77 & 34.21 & 45.51 & 27.68 & 25.55 & 92.20 & 68.20 & 68.00 \\
 & \textsc{ard + prm} & 85.80 & 53.36 & 49.33 & 59.89 & 30.32 & 28.81 & 95.20 & 73.00 & 70.60 \\
 & \textbf{\textsc{safer}} & \textbf{88.91} & \textbf{56.21} & \textbf{51.51} & \textbf{62.23} & \textbf{31.42} & \textbf{30.02} & \textbf{96.22} & \textbf{77.02} & \textbf{72.24} \\
\midrule
\textbf{Swin-Ti}~\cite{liu2021swin} & \textsc{pgd-at (sgd)} & 79.34 & 47.95 & 45.98 & 56.46 & 28.42 & 22.10 & 94.80 & 72.80 & 71.80 \\
 & \textsc{ard + prm} & 82.63 & 48.87 & 45.31 & 58.12 & 30.32 & 24.35 & 96.20 & 74.40 & 71.20 \\
 & \textbf{\textsc{safer}} & \textbf{84.15} & \textbf{50.99} & \textbf{49.01} & \textbf{60.11} & \textbf{31.45} & \textbf{25.51} & \textbf{97.33} & \textbf{75.45} & \textbf{72.69} \\
\midrule
\textbf{Swin-S} & \textsc{pgd-at (sgd)} & 79.34 & 48.53 & 44.88 & 57.89 & 29.39 & 23.94 & 95.40 & 74.00 & 73.80 \\
 & \textsc{pgd-at (sam)} & 82.95 & 51.60 & 47.25 & 58.40 & 30.05 & 25.00 & 95.85 & 74.65 & 76.10 \\
 & \textsc{ard + prm} & 84.46 & 50.02 & 46.17 & 59.12 & 30.23 & 25.21 & 96.00 & 75.00 & 74.80 \\
 & \textbf{\textsc{safer}} & \textbf{86.52} & \textbf{52.00} & \textbf{50.10} & \textbf{61.78} & \textbf{31.97} & \textbf{26.40} & \textbf{97.10} & \textbf{78.76} & \textbf{78.00} \\
\midrule

\textbf{Swin-B} & \textsc{pgd-at (sgd)} & 83.36 & 50.19 & 46.89 & 57.88 & 29.11 & 24.64 & 96.40 & 75.80 & 74.60 \\
& \textsc{ard + prm} & 84.16 & 51.47 & 47.50 & 59.94 & 30.34 & 26.00 & 97.20 & 77.40 & 76.20 \\
& \textsc{\citet{tian2022deeper}} & 84.50 & 52.42 & 50.10 & 58.67 & 29.34 & 25.98 & 89.88 & 75.45 & 71.18 \\
 & \textsc{\cite{tian2022deeper} + safer} & 85.55 & 53.01 & 51.11 & 59.78 & 31.00 & 27.22 & 91.48 & 76.97 & 73.20 \\
 & \textbf{\textsc{safer}} & \textbf{86.78} & \textbf{53.65} & \textbf{52.00} & \textbf{61.20} & \textbf{32.91} & \textbf{27.13} & \textbf{98.45} & \textbf{80.13} & \textbf{77.68} \\
\bottomrule
\end{tabular}
\end{table*}

\begin{table*}[htbp]
\caption{PGD-20 Adversarial Accuracies on CIFAR-10 for Different Training/Finetuning Methods with Differences from SAFER}
\label{tab:ft_comp}
\centering
\footnotesize
\begin{tabular}{l@{\hspace{10pt}}c@{\hspace{15pt}}c@{\hspace{10pt}}c@{\hspace{15pt}}c@{\hspace{10pt}}c@{\hspace{15pt}}c}
\toprule
\textbf{Model} & \textsc{PGD-AT (SGD)} & \textsc{RiFT} & \textsc{Clat} & \textbf{SAFER} & \textsc{SAFER - RiFT} \(\Delta\) & \textsc{SAFER - Clat} \(\Delta\) \\
\midrule
\textbf{WRN-34-10}~\cite{zagoruyko2017wide} & 57.40 & 55.01 & 57.11 & \textbf{59.02} & \textcolor{ForestGreen}{+\(4.01\)} & \textcolor{ForestGreen}{+\(1.91\)} \\
\textbf{RN-18}~\cite{he2016deep} & 53.63 & 54.65 & 55.37 & \textbf{56.98} & \textcolor{ForestGreen}{+\(2.33\)} & \textcolor{ForestGreen}{+\(1.61\)} \\
\hline
\textbf{DeiT-Ti} & 48.10 & 44.32 & 46.29 & \textbf{68.50} & \textcolor{ForestGreen}{+\(24.18\)} & \textcolor{ForestGreen}{+\(22.21\)} \\
\textbf{ViT-S} & 51.88 & 45.93 & 46.12 & \textbf{68.89} & \textcolor{ForestGreen}{+\(22.96\)} & \textcolor{ForestGreen}{+\(22.77\)} \\
\bottomrule
\end{tabular}
\end{table*}

\subsubsection{Whitebox Robustness}  
Table~\ref{tab:wb} demonstrates SAFER's effectiveness in enhancing both clean and adversarial accuracies across diverse transformer architectures and attack types in a white-box setting. The results underscore SAFER’s robust performance across datasets and training methods, including over SOTA techniques (for example, ARD + PRM ~\cite{mo2022adversarialtrainingmeetsvision} and~\citet{tian2022deeper}) from RobustBench. The consistency across different architectures and model sizes further illustrates SAFER’s scalability, even in larger models.
Notably, models trained with SAFER retain robustness against Auto Attacks, despite having been trained solely on PGD attacks. This resilience suggests that SAFER further reduces the overfitting on the specific attack model is trained on.

As we utilize SAM techniques in SAFER optimization, we highlight the comparison between SAFER and naively applying SAM in the PGD-AT optimization process. As shown in Table~\ref{tab:wb}, though SAM helps the clean accuracy and robustness over SGD in PGD-AT, the performance improvement is limited due to the complexity in convergence, such that SAM becomes less effective on larger models. In comparison, SAFER resolves the convergence complexity by only targeting the layers that are the most prone to overfitting, significantly enhancing the final performance. 

Due to space limitations, only PGD-20 and AA are used for robustness evaluation in Table~\ref{tab:wb}. We show the validity of our attack convergence and the consistency of SAFER's robustness across different attacks in Appendix D.

\paragraph{Benchmarking SAFER Against Existing Layer-selective Fine-Tuning Approaches}
Furthermore, we highlight the limitations of existing layer-selective fine-tuning techniques, like CLAT~\cite{clat} and RiFT~\cite{rift}, when applied to vision transformers. As shown in \cref{tab:ft_comp}, while previous methods improve adversarial accuracy in CNN models such as Wide ResNet-50 and ResNet-18, they noticeably reduce adversarial accuracy in ViTs. This suggests that the feature-based layer selection metrics designed for CNNs are not effective for ViTs due to the diverse layer types and complicated architectures. In contrast, SAFER demonstrates superior performance, clearly outperforming these techniques.

\begin{table*}[htbp]
\caption{Comparative Analysis of Black-box Auto Attack Accuracy on CIFAR-10 and Imagenette. Each row is the attacker, and each column is the victim.}
\label{tab:bb_autoattack}
\centering
\footnotesize
\begin{tabular}{l@{\hspace{5pt}}l@{\hspace{10pt}}c@{\hspace{15pt}}c@{\hspace{15pt}}c@{\hspace{15pt}}c@{\hspace{15pt}}c@{\hspace{15pt}}c@{\hspace{15pt}}c@{\hspace{15pt}}c}
\toprule
\textbf{Network} & \textbf{Method} & \multicolumn{4}{c}{\textsc{Cifar-10 Adv. Acc. (\%)}} & \multicolumn{4}{c}{\textsc{Imagenette Adv. Acc. (\%)}} \\
\cmidrule(lr){3-6} \cmidrule(lr){7-10}
 &  & \textbf{DeiT-Ti} & \textbf{DeiT-S} & \textbf{ViT-S} & \textbf{Swin-B} & \textbf{DeiT-Ti} & \textbf{DeiT-S} & \textbf{ViT-S} & \textbf{Swin-B} \\
\midrule
\textbf{DeiT-Ti} & \textsc{pgd-at (sgd)} & - & 50.22 & 50.15 & 53.22 & - & 70.45 & 69.21 & 80.21 \\
 & \textsc{safer} & - & 54.45 & 53.86 & 57.89 & - & 74.89 & 72.13 & 84.93 \\
\midrule
\textbf{DeiT-S} & \textsc{pgd-at (sgd)} & 50.71 & - & 49.83 & 51.22 & 67.67 & - & 70.45 & 83.48 \\
 & \textsc{safer} & 52.28 & - & 55.01 & 55.43 & 71.50 & - & 73.68 & 86.72 \\
\midrule
\textbf{ViT-S} & \textsc{pgd-at (sgd)} & 50.13 & 50.67 & - & 50.41 & 68.89 & 71.28 & - & 83.91 \\
 & \textsc{safer} & 53.91 & 55.16 & - & 54.82 & 70.23 & 75.97 & - & 87.43 \\
\midrule
\textbf{Swin-B} & \textsc{pgd-at} & 52.01 & 52.45 & 54.03 & - & 68.23 & 76.00 & 74.88 & - \\
 & \textsc{safer} & 55.91 & 57.01 & 56.65 & - & 72.45 & 79.21 & 76.53 & - \\
\bottomrule
\end{tabular}
\end{table*}

\begin{table*}[htbp]
\caption{Comparative Analysis of Black-box PGD-20 Accuracy on CIFAR-10 and Imagenette. Each row is the attacker, and each column is the victim.}
\label{tab:bb_pgd}
\centering
\footnotesize
\begin{tabular}{l@{\hspace{5pt}}l@{\hspace{10pt}}c@{\hspace{15pt}}c@{\hspace{15pt}}c@{\hspace{15pt}}c@{\hspace{15pt}}c@{\hspace{15pt}}c@{\hspace{15pt}}c@{\hspace{15pt}}c}
\toprule
\textbf{Network} & \textbf{Method} & \multicolumn{4}{c}{\textsc{Cifar-10 Adv. Acc. (\%)}} & \multicolumn{4}{c}{\textsc{Imagenette Adv. Acc. (\%)}} \\
\cmidrule(lr){3-6} \cmidrule(lr){7-10}
 &  & \textbf{DeiT-Ti} & \textbf{DeiT-S} & \textbf{ViT-S} & \textbf{Swin-B} & \textbf{DeiT-Ti} & \textbf{DeiT-S} & \textbf{ViT-S} & \textbf{Swin-B} \\
\midrule
\textbf{DeiT-Ti} & \textsc{pgd-at (sam)} & - & 57.50 & 60.30 & 54.10 & - & 69.70 & 70.01 & 81.32 \\
 & \textsc{safer} & - & 74.60 & 68.40 & 58.20 & - & 72.53 & 72.24 & 84.45 \\
\midrule
\textbf{DeiT-S} & \textsc{pgd-at (sam)} & 54.50 & - & 59.00 & 53.50 & 70.20 & - & 71.45 & 80.73 \\
 & \textsc{safer} & 71.80 & - & 64.23 & 57.80 & 70.88 & - & 73.89 & 85.01 \\
\midrule
\textbf{ViT-S} & \textsc{pgd-at (sam)} & 55.50 & 58.00 & - & 55.80 & 68.45 & 71.48 & - & 79.45 \\
 & \textsc{safer} & 73.98 & 76.50 & - & 60.20 & 71.87 & 73.96 & - & 83.20 \\
\midrule
\textbf{Swin-B} & \textsc{pgd-at (sam)} & 58.10 & 59.80 & 64.30 & - & 72.76 & 74.59 & 74.88 & - \\
 & \textsc{safer} & 76.21 & 78.90 & 68.50 & - & 74.53 & 75.82 & 76.92 & - \\
\bottomrule
\end{tabular}
\end{table*}

\subsubsection{Blackbox Robustness}
In addition to white-box results, Tables~\ref{tab:bb_autoattack} and~\ref{tab:bb_pgd} provide black-box robustness evaluations (Auto Attack and PGD, respectively), comparing models trained solely with PGD-AT versus those enhanced with SAFER, using the same attack settings as in white-box evaluations.

As a sanity check, the higher accuracies observed in black-box settings compared to white-box settings indicate that gradient masking is not present in models using SAFER, confirming the reliability of our white-box robustness evaluation.

In line with white-box results, models trained with SAFER also outperform those trained solely with PGD-AT in black-box settings, demonstrating greater resilience across both black-box and white-box settings, regardless of attack method or model architecture. These findings underscore SAFER as a robust, adaptable solution for adversarial training across diverse configurations and attack scenarios.

\subsubsection{PEFT methods}

\begin{table}[htbp]
\caption{Effect of SAFER on PEFT Models (LoRA/DoRA) for CIFAR-10 and Imagenette with PGD-20 Adversarial Accuracy.}
\label{tab:peft_comp}
\centering
\footnotesize
\resizebox{\linewidth}{!}{
\begingroup
    \setlength{\tabcolsep}{3pt}
\begin{tabular}{llcccc}
\toprule
\textbf{Network} & \textbf{Method} & \multicolumn{2}{c}{\textsc{Cifar-10}} & \multicolumn{2}{c}{\textsc{Imagenette}} \\
\cmidrule(lr){3-4} \cmidrule(lr){5-6}
 &  & \textbf{Clean} & \textbf{Adv. Acc.} & \textbf{Clean} & \textbf{Adv. Acc.} \\
\midrule
\textbf{DeiT-Ti} & \textsc{Lora PGD-AT (SGD)} & 57.01 & 40.23 & 58.30 & 41.10 \\
 & \textsc{Lora PGD-AT (SAM)} & 71.45 & 51.45 & 73.20 & 52.30 \\
 & \textsc{Lora SAFER} & \textbf{78.12} & \textbf{63.50} & \textbf{80.50} & \textbf{64.20} \\
 \cmidrule{2-6}
 & \textsc{Dora PGD-AT (SGD)} & 59.65 & 42.86 & 60.10 & 43.00 \\
 & \textsc{Dora PGD-AT (SAM)} & 73.30 & 53.82 & 74.00 & 54.10 \\
 & \textsc{Dora SAFER} & \textbf{80.18} & \textbf{65.55} & \textbf{81.10} & \textbf{65.80} \\
\midrule
\textbf{ViT-B} & \textsc{Lora PGD-AT (SGD)} & 75.50 & 60.70 & 77.00 & 62.10 \\
 & \textsc{Lora PGD-AT (SAM)} & 78.80 & 65.10 & 81.30 & 66.80 \\
 & \textsc{Lora SAFER} & \textbf{81.40} & \textbf{70.20} & \textbf{84.90} & \textbf{69.10} \\
 \cmidrule{2-6}
 & \textsc{Dora PGD-AT (SGD)} & 76.80 & 61.50 & 78.50 & 63.00 \\
 & \textsc{Dora PGD-AT (SAM)} & 80.90 & 66.10 & 82.20 & 68.00 \\
 & \textsc{Dora SAFER} & \textbf{81.99} & \textbf{71.30} & \textbf{85.30} & \textbf{71.20} \\ 
\bottomrule
\end{tabular}
\endgroup
}
\end{table}

With the recent advancements in large language models, PEFT methods such as LoRA~\cite{hu2021loralowrankadaptationlarge} and DORA~\cite{dora} have become popular approaches for updating weights when tuning large transformer models. To evaluate the generalizability of our method, we provide results for models trained using SAFER and compare them with models trained with PGD-AT, using both SGD and SAM optimizers. PEFT methods are applied for weight updates.The results, presented in Table~\ref{tab:peft_comp}, reveal a consistent performance trend similar to that of full fine-tuning, where SAFER consistently outperforms across different models, datasets, and PEFT techniques.

\subsection{Ablation Studies}
\label{sec:ablation}

\begin{figure*}[ht]
    \centering
    \includegraphics[width=.95\textwidth]{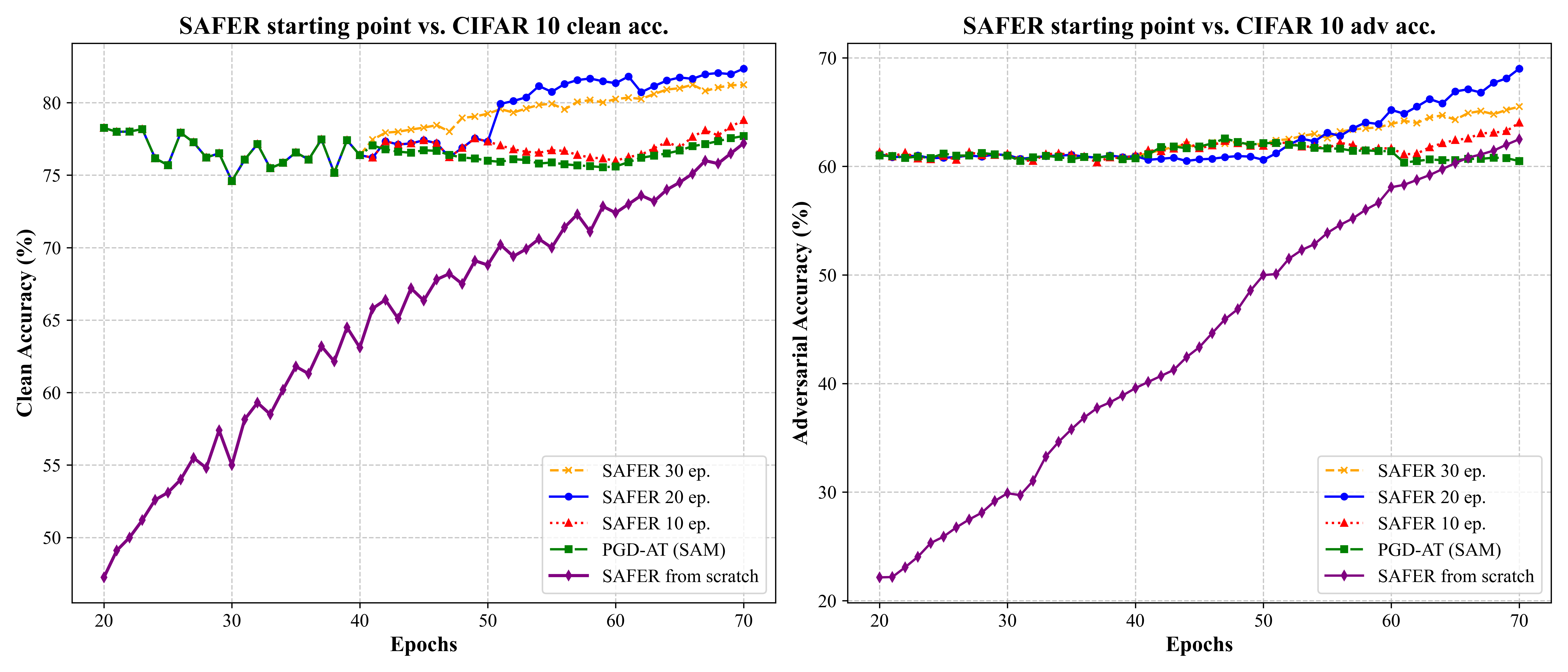}  
    \caption{
    SAFER performance at different starting points on CIFAR-10 with DeiT-Ti: clean (left) vs. adversarial (right) accuracies.}
    \label{fig:eps}
\end{figure*}

\subsubsection{Pretraining epochs}
As discussed in~\cref{method}, we apply SAFER after an initial phase of adversarial training. Here, we examine how varying the number of pretraining epochs affects performance. Figure \ref{fig:eps} presents training curves for different allocations of PGD pretraining and fine-tuning epochs within a 70-epoch training budget.
In models trained solely with PGD-AT, even with SAM optimizer being applied, both clean and adversarial accuracy graphs reveal limitations. In the clean accuracy graph, PGD-AT accuracy plateaus and then declines slightly, signaling overfitting, as documented in previous research \citep{rice2020overfitting}. By contrast, SAFER continues to improve clean accuracy, effectively mitigating this overfitting issue. In the adversarial accuracy graph, PGD-AT performance falls significantly short of SAFER’s, highlighting a notable gap in robustness.
Notably, incorporating SAFER at any stage of training results in higher clean accuracy and robustness at convergence. 

We also include results for applying SAFER from scratch (0 epochs of pretraining) with some interesting observations. Although the model trained exclusively with SAFER eventually achieves competitive performance with PGD-AT, its convergence is significantly slower. This suggests that full model training is beneficial in the early stages for rapid convergence, while layer-selective training helps mitigate overfitting and boosts model performance in later stages.

\subsubsection{Layer Selection}

SAFER’s effectiveness hinges on selecting the layers most prone to overfitting for fine-tuning, while keeping the remaining layers frozen. To verify the importance of this selection, we compare SAFER with an alternative approach where randomly chosen layers are dynamically fine-tuned instead of those identified by our method. Results of this comparison are presented in Table \ref{tab:ablation_1}, demonstrating that targeting the layers identified by SAFER significantly improves both adversarial robustness and clean accuracy.
Furthermore, as shown in Appendix A, certain layers are consistently selected within the same model architecture, even across diverse datasets. This consistency suggests that some layers possess inherent properties that predispose them to overfit, reinforcing the importance of accurately identifying and targeting these layers for SAFER to be effective.

Additional ablation studies in the Appendix B and C examine two key factors in SAFER’s performance: frequency of dynamic layer selection and the number of layers chosen for fine-tuning. Results indicate that dynamically re-evaluating layers for fine-tuning is crucial, as new overfitting layers can emerge once previously selected layers shift away from overfitting. Additionally, the number of layers selected is crucial, as choosing too many layers complicates the optimization process and can result in poorer performance. The hyperparameters used in our main experiments are informed by these observations.

\begin{table}[tb]
\centering
\footnotesize
\caption{PGD-20 and Auto Attack (AA) adversarial accuracies for models with sharpness-selected vs. randomly selected layers for fine-tuning in SAFER.}
\label{tab:ablation_1}
\vspace{5pt}
\resizebox{\linewidth}{!}{
\begingroup
    \setlength{\tabcolsep}{3pt}
\begin{tabular}{lcccccc}
\toprule
Network & \multicolumn{3}{c}{Sharpness-Selected Layers} & \multicolumn{3}{c}{Randomly Selected Layers} \\
\cmidrule(lr){2-4} \cmidrule(lr){5-7}
& Clean Acc. & PGD-20 & AA & Clean Acc. & PGD-20 & AA \\
\midrule
DeiT-Ti & 82.36 & 68.50 & 50.12 & 79.15 & 65.46 & 45.38 \\
ViT-S & 83.40 & 68.89 & 50.12 & 80.19 & 64.35 & 46.19 \\
Swin-B & 86.78 & 53.65 & 52.00 & 81.73 & 49.72 & 44.38 \\
\bottomrule
\end{tabular}
\endgroup
}
\end{table}

\subsection{Overhead analysis}


Lastly, we show that the time required to determine layer sharpness is negligible within the overall adversarial training process. As shown in ~\cref{tab:layer_time}, layer sharpness ranking can be reliably computed with a batch size as small as 50, with top-ranking layers consistent across larger batches. We conducted over 1,000 runs per network, randomly selecting data for sharpness estimation, and found remarkable consistency in the computed layer rankings. This stability allows us to use only \textbf{0.001\%} of the training data for sharpness estimation every 10 epochs, adding just \textbf{0.2\%} extra time to the standard adversarial training process, as shown in ~\cref{tab:layer_time}. For reference, one epoch of DeiT-Ti PGD-AT with SGD on CIFAR-10 takes approximately 290 seconds. Although an additional backpropagation step is required for SAM compared to SGD, the overhead of using SAM-based SAFER remains minimal. A SAFER fine-tuning epoch takes 298 seconds, adding only $\sim3\%$ more time to SGD. This is because computing adversarial examples, which requires multiple backpropagations, takes up the majority of the time.


\begin{table}[tb]
\centering
\caption{DeiT-Ti fine-tuning layers selected by SAFER with varying data amounts and corresponding computation times.}

\label{tab:layer_time}
\footnotesize 
\resizebox{\linewidth}{!}{
\begingroup
    \setlength{\tabcolsep}{2pt}  
\begin{tabular}{ccccc}
\toprule
\textsc{Batch Size} & \multicolumn{2}{c}{\textsc{CIFAR-10}} & \multicolumn{2}{c}{\textsc{Imagenette}} \\
\cmidrule(lr){2-3} \cmidrule(lr){4-5}
& \textsc{Critical Layers} & \textsc{Time (s)} & \textsc{Critical Layers} & \textsc{Time (s)} \\
\midrule
50 & 11, 10, 13, 8, 16 & 7.20 & 11, 10, 13, 8, 16 & 9.13 \\
100 & 11, 10, 13, 8, 16 & 8.50 & 11, 10, 13, 9, 16 & 10.20 \\
200 & 11, 10, 13, 8, 14 & 13.00 & 11, 10, 13, 8, 16 & 15.45 \\
300 & 11, 10, 13, 8, 16 & 16.15 & 11, 10, 13, 9, 16 & 17.60 \\
500 & 11, 10, 16, 13, 14 & 17.34 & 11, 10, 13, 8, 16 & 19.50 \\
\bottomrule
\end{tabular}
\endgroup
}
\end{table}

\section{Conclusions}

This work introduces SAFER, a layer-selective fine-tuning framework for Vision Transformers (ViTs) that addresses adversarial overfitting by selectively refining layers identified as most prone to adversarial overfitting using sharpness-aware minimization (SAM). Our results show that fine-tuning a limited subset of layers achieves notable improvements in both clean and adversarial accuracy across various architectures and baseline adversarial training methods. Additionally, we demonstrate that SAFER integrates effectively with Parameter-Efficient Fine-Tuning (PEFT) approaches, underscoring its versatility in transformer-based models.
We limit the scope of this work to enhancing empirical robustness in ViTs. Open questions remain regarding why certain layers in ViTs become prone to overfitting, how they might be identified with greater precision, and whether architectural or training modifications could further improve robustness. A deeper theoretical exploration of these questions is left for future work.

%% file: appendix.tex

\setcounter{table}{7}  
\setcounter{figure}{2}  

\maketitlesupplementary  
\appendix  

\section{Top-sharpness layers by model and dataset}
\label{sec:ci_m_d}

Table~\ref{ap:tab:ci} lists the indices of layers exhibiting the highest sharpness in a PGD-AT pretrained ViT model. Notably, while sharp layers vary across models, the dataset used for training and evaluation has minimal effect on the sharpness ranking of layers. This aligns with our observation in Table~7, where sharpness rankings remain consistent across different random evaluation batches. This consistency suggests that certain layers have inherent structural properties, predisposing them to overfitting during adversarial training. A theoretical exploration of this observation is reserved for future work. Theoretical analysis on this observation is left for future work.

\begin{table}[htb]
\centering
\small
\caption{Indices of the Top-5 sharpest layers by model and dataset. Layers selected in SAFER finetuning are bolded.}
\label{ap:tab:ci}
\begingroup
    \setlength{\tabcolsep}{4pt}  
{\scshape  
\begin{tabular}{@{}lcc@{}}  
\toprule
\textsc{Model} & \textsc{CIFAR10} & \textsc{CIFAR100} \\
\midrule
DeiT-Ti & \textbf{11, 10}, 13, 8, 16 & \textbf{11, 10}, 13, 8, 16 \\
ViT-S &\textbf{5, 9}, 14, 20, 25 &  \textbf{5, 9}, 15, 20, 25 \\
Swin-B & \textbf{4, 15, 32, 46}, 50 & \textbf{4, 16, 32, 46}, 50 \\
\bottomrule
\end{tabular}
}
\endgroup
\end{table}

\section{Ablation: Dynamic layer selection}
\label{sec:dyn}

During the SAFER finetuning process, we recompute the sharpness measurement every 10 epochs to update our selection of layers most susceptible to overfitting. Table~\ref{tab:ablation_cifar10_fixed_critical} shows that dynamically selecting layers for finetuning is crucial to SAFER’s performance. In contrast, fixing the same set of layers that are selected on the initial pretrained model throughout finetuning results in significantly lower accuracies compared to the baselines. Finetuning in this study was conducted for 20 epochs. Although not shown, the performance gap becomes more pronounced with extended finetuning, as the initially selected layers become less prone to overfitting, providing minimal improvements with further tuning.

\begin{table}[htb]
\centering
\footnotesize
\caption{Ablation study on dynamic and fixed sharp layer selection for SAFER on CIFAR-10: The columns present clean, PGD-20 and Auto Attack (AA) evaluation accuracies for models trained with SAFER, using dynamic or fixed layers for fine-tuning.}
\label{tab:ablation_cifar10_fixed_critical}
\centering
\small
\begingroup
    \setlength{\tabcolsep}{2pt}
{\scshape  
\begin{tabular}{lcccccc}
\toprule
Network & \multicolumn{3}{c}{Dynamic Layers} & \multicolumn{3}{c}{Fixed Layers} \\
\cmidrule(lr){2-4} \cmidrule(lr){5-7}
& Clean & PGD-20 & AA & Clean & PGD-20 & AA \\
\midrule
DeiT-Ti & 82.36 & 68.50 & 50.12 & 80.10 & 64.49 & 47.53 \\
ViT-S   & 83.40 & 68.89 & 50.12 & 79.92 & 63.12 & 46.21 \\
Swin-B  & 86.52 & 53.65 & 52.00 & 84.39 & 50.45 & 50.19 \\
\bottomrule
\end{tabular}
}
\endgroup
\end{table}

\section{Ablation: Number of layers chosen}

\begin{figure*}[htb]
    \centering
    \includegraphics[width=\textwidth]{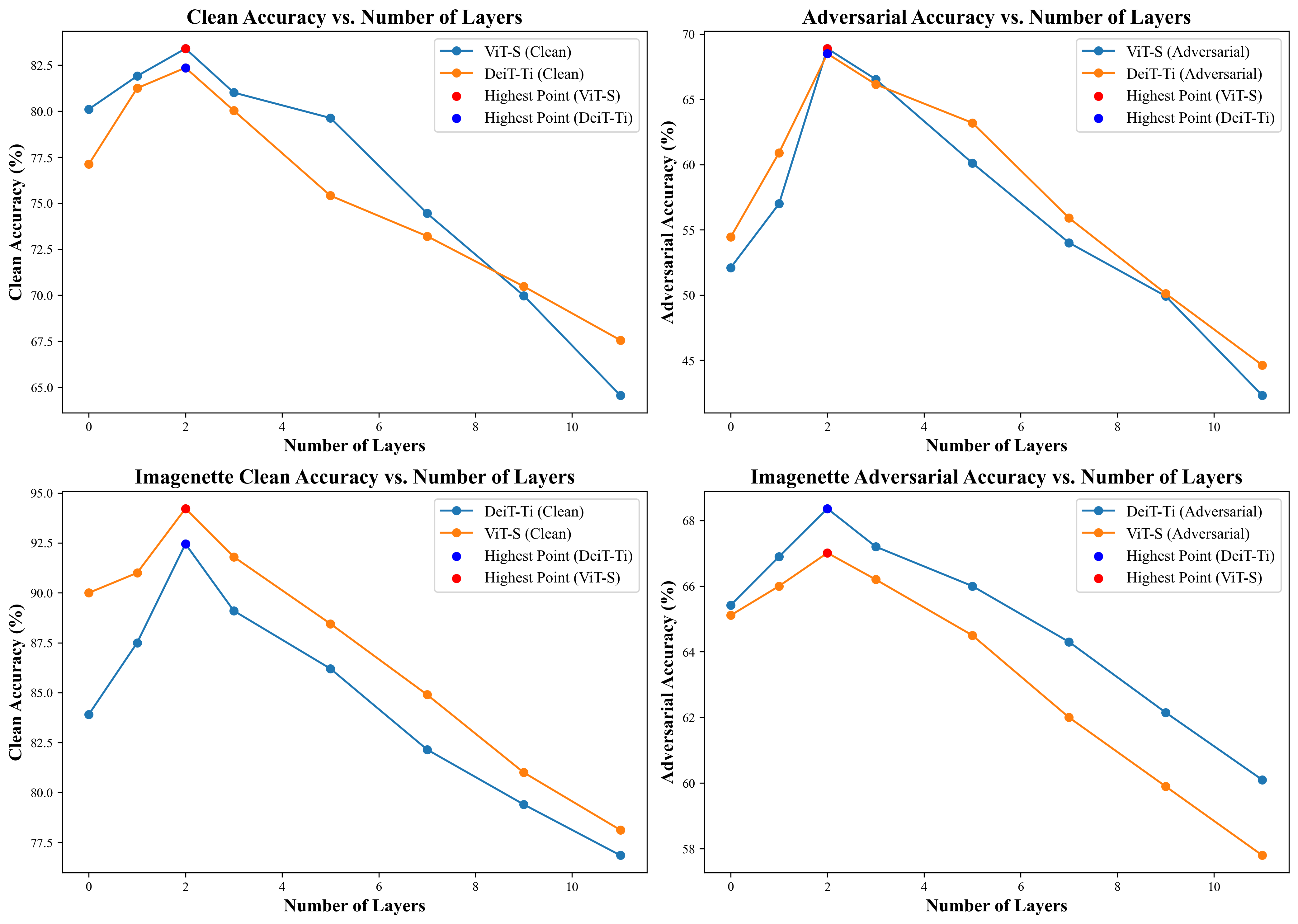}
    \caption{Performance comparison of DeiT-Ti and ViT-S as a function of number of sharp layers selected for SAFER finetuning. 
    The top row shows CIFAR-10 clean and adversarial accuracy, while the bottom row shows Imagenette results. 
    The number ``0'' on the X axis corresponds to PGD-AT (SAM) without SAFER, where fine-tuning is performed on the entire model. 
    The highest performance points are highlighted in blue for DeiT-Ti and red for ViT-S.}

    \label{fig:accuracy_layers}
\end{figure*}

Figure~\ref{fig:accuracy_layers} illustrates the performance variation between adversarial accuracy and the number of layers selected for SAFER finetuning in both DeiT-Ti and ViT-S. Initially, increasing the number of layers improves model flexibility, resulting in enhanced SAFER performance. However, beyond a certain point, finetuning additional layers reduces SAFER’s effectiveness, likely due to the model focusing on less-relevant layers, which complicates convergence under the SAM objective. This trend is consistent across datasets, with results shown for Imagenette and CIFAR-10.

Both models identify selecting the Top-2 layers (approximately 5\% of the 36 total layer options) as leading to the best results. As ViT-S layers are significantly larger, selecting additional layers for ViT-S results in a steeper decline in performance, primarily due to optimization difficulties arising from the increased parameter count. This observation highlights the importance of selecting the optimal number of layers during SAFER finetuning.


\section{Additional attacks for robustness evaluation}
\label{sec:strength}

\subsection{Convergence of the PGD attack}

In the main paper, we report PGD attack robustness using attacks with 20 gradient ascent steps. As suggested by~\citet{carlini2019evaluating}, PGD attacks with insufficient update steps may be ineffective due to gradient masking, resulting in inaccurate robustness measurements. To address this, \cref{tab:pgd_steps} presents the robustness results under PGD attacks with increased steps for selected models reported in Table 1. As shown in the table, increasing the attack steps does not result in further decreases in model robustness. This demonstrates that the adversarial images generated in the main paper originate from well-converged attacks, and adding more steps does not improve convergence.

\begin{table}[htb]
\centering
\caption{SAFER adversarial accuracy on CIFAR-10 under PGD attacks with 20, 50, and 100 steps}
\label{tab:pgd_steps}
\begin{sc}
\small
\begin{tabular}{lccc}
\toprule
\textbf{model} & \textbf{pgd-20 (\%)} & \textbf{pgd-50 (\%)} & \textbf{pgd-100 (\%)} \\
\midrule
\textsc{deit-ti}    & 68.50 & 68.12 & 68.04 \\
\textsc{vit-s}      & 68.89 & 68.51 & 68.73 \\
\textsc{convit-b}   & 56.21 & 56.34 & 56.22 \\
\textsc{swin-b}     & 53.65 & 53.22 & 53.05 \\
\bottomrule
\end{tabular}
\end{sc}
\end{table}

\subsection{Additional attack types}
To further demonstrate SAFER's effectiveness in improving model robustness, we compare models trained with SAFER to those trained with PGD-AT (SAM) on the CIFAR-10 dataset, as reported in Table~1. The evaluation includes stronger white-box attacks that are not limited to $\ell_\infty$-bounded constraints. Table~\ref{ap:tab:attack} shows the robustness results under the FAB attack~\cite{croce2020minimally}, StAdv attack~\cite{xiao2018spatially}, PIXEL attack~\cite{su2019one}, $\ell_\infty$-bounded PGD attacks with higher strengths, and $\ell_2$-bounded PGD attack. Across all evaluated attacks, SAFER-trained models consistently demonstrate robustness improvements over baseline models.

\begin{table}[htb]
\caption{Adversarial accuracies across various attacks on CIFAR-10, comparing models trained on DeiT-Ti and ViT-S without and with SAFER training/finetuning, respectively. Positive blue values indicate performance improvements achieved with SAFER-trained models over PGD-AT (SAM) baseline.}
\label{ap:tab:attack}
\begin{center}
\small
\begin{sc}
\begingroup
    \setlength{\tabcolsep}{2pt}
\begin{tabular}{llcc}
\toprule
Attack & Method & DeiT-Ti (\%) & ViT-S (\%) \\
\midrule
FAB & PGD-AT (SAM) & 24.79 & 26.52 \\
    & SAFER & {\color{blue}+3.36} & {\color{blue}+2.61} \\
\midrule
StAdv & PGD-AT (SAM) & 19.60 & 20.21 \\
      & SAFER & {\color{blue}+3.85} & {\color{blue}+4.54} \\
\midrule
Pixel & PGD-AT (SAM) & 7.30 & 8.40 \\
      & SAFER & {\color{blue}+1.40} & {\color{blue}+1.50} \\
\midrule
PGD-20 \( L_\infty \) & PGD-AT (SAM) & 54.45 & 52.10 \\
($\epsilon = 0.03$) & SAFER & {\color{blue}+14.05} & {\color{blue}+16.79} \\
\midrule
PGD-20 \( L_\infty \) & PGD-AT (SAM) & 47.20 & 46.28 \\
($\epsilon = 0.05$) & SAFER & {\color{blue}+6.59} & {\color{blue}+8.63} \\
\midrule
PGD-20 \( L_\infty \) & PGD-AT (SAM) & 40.25 & 41.79 \\
($\epsilon = 0.07$) & SAFER & {\color{blue}+9.94} & {\color{blue}+10.03} \\
\midrule
PGD-20 \( L_2 \) & PGD-AT (SAM) & 56.79 & 56.05 \\
($\epsilon = 0.03$) & SAFER & {\color{blue}+12.33} & {\color{blue}+14.13} \\
\bottomrule
\end{tabular}
\endgroup
\end{sc}
\end{center}
\end{table}

\section{Learning curve under extended training}
\begin{figure*}[htb]
    \centering
    \includegraphics[width=\textwidth]{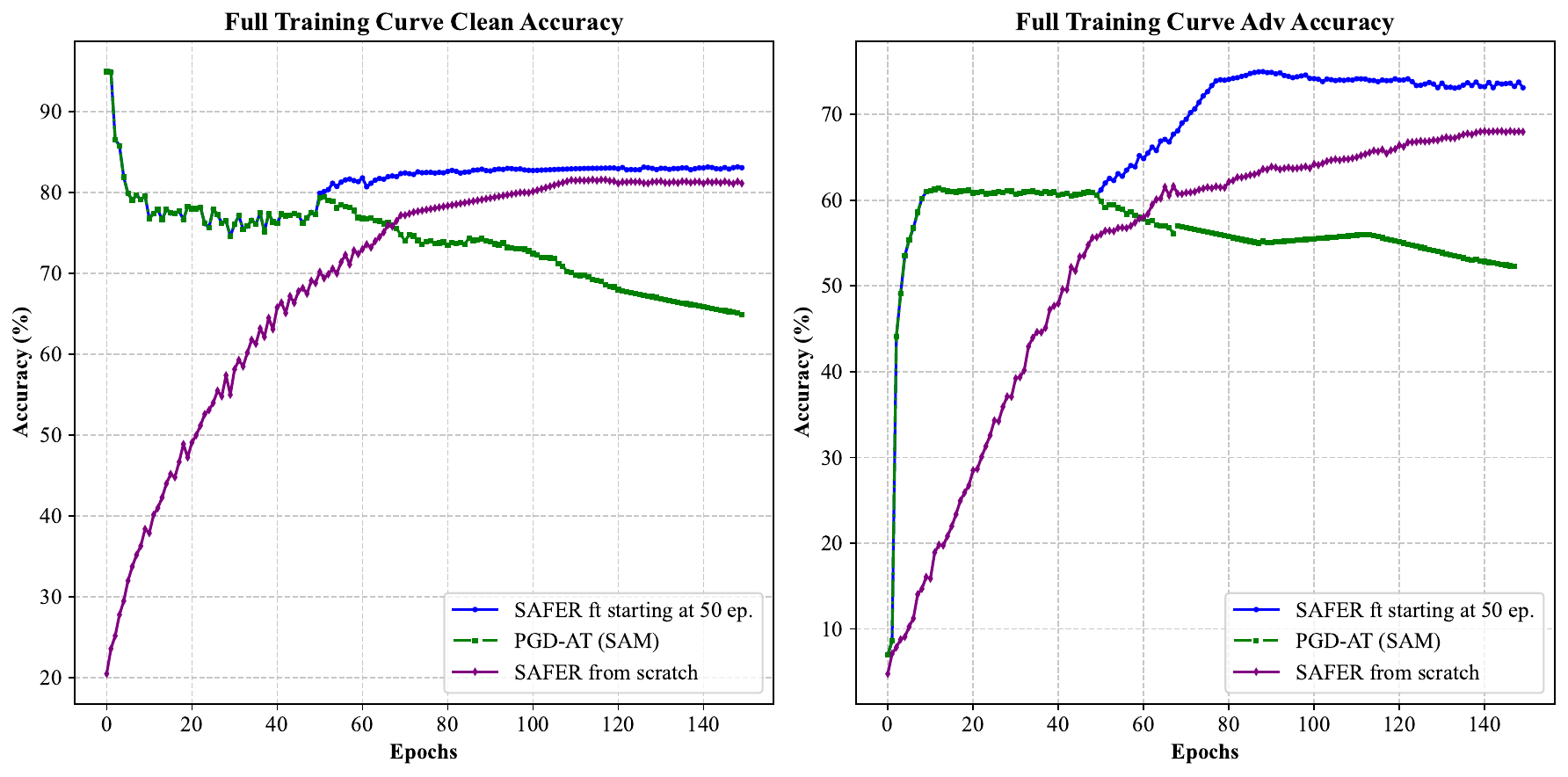}
    \caption{SAFER vs. PGD-AT (SAM) performance on CIFAR-10 with DeiT-Ti: clean (left) vs. adversarial (right) accuracies.}
    \label{fig:full_curve}
\end{figure*}

It has been observed that adversarial overfitting can be mitigated by early stopping~\cite{rice2020overfitting}. SAFER is designed to eliminate the need for early stopping and fully leverage the model's learning potential throughout the finetuning process. To this end, we extend the learning curve experiments in Figure 2 to 150 adversarial training epochs. A cosine learning rate scheduler is used so that the learning rate decays to 0 by epoch 150. 
As shown in Figure~\ref{fig:full_curve}, even with the SAM optimizer, models trained with PGD-AT show a consistent decline in performance with additional epochs of adversarial training, highlighting the effects of overfitting.
In contrast, models trained with SAFER (whether pretrained model or from scratch), show consistent performance and robustness improvement throughout all epochs. This further proves the effectiveness of SAFER in countering overfitting.  

%% file: cvpr.bbl
\begin{thebibliography}{40}
\providecommand{\natexlab}[1]{#1}
\providecommand{\url}[1]{\texttt{#1}}
\expandafter\ifx\csname urlstyle\endcsname\relax
  \providecommand{\doi}[1]{doi: #1}\else
  \providecommand{\doi}{doi: \begingroup \urlstyle{rm}\Url}\fi

\bibitem[Athalye et~al.(2018)Athalye, Carlini, and Wagner]{athalye2018obfuscated}
Anish Athalye, Nicholas Carlini, and David Wagner.
\newblock Obfuscated gradients give a false sense of security: Circumventing defenses to adversarial examples, 2018.

\bibitem[Bai et~al.(2021)Bai, Mei, Yuille, and Xie]{bai2021transformersrobustcnns}
Yutong Bai, Jieru Mei, Alan Yuille, and Cihang Xie.
\newblock Are transformers more robust than cnns?, 2021.

\bibitem[Carion et~al.(2020)Carion, Massa, Synnaeve, Usunier, Kirillov, and Zagoruyko]{carion2020end}
Nicolas Carion, Francisco Massa, Gabriel Synnaeve, Nicolas Usunier, Alexander Kirillov, and Sergey Zagoruyko.
\newblock End-to-end object detection with transformers.
\newblock In \emph{European conference on computer vision}, pages 213--229. Springer, 2020.

\bibitem[Carlini and Wagner(2017)]{carlini2017evaluating}
Nicholas Carlini and David Wagner.
\newblock Towards evaluating the robustness of neural networks, 2017.

\bibitem[Carlini et~al.(2019)Carlini, Athalye, Papernot, Brendel, Rauber, Tsipras, Goodfellow, Madry, and Kurakin]{carlini2019evaluating}
Nicholas Carlini, Anish Athalye, Nicolas Papernot, Wieland Brendel, Jonas Rauber, Dimitris Tsipras, Ian Goodfellow, Aleksander Madry, and Alexey Kurakin.
\newblock On evaluating adversarial robustness.
\newblock \emph{arXiv preprint arXiv:1902.06705}, 2019.

\bibitem[Carmon et~al.(2022)Carmon, Raghunathan, Schmidt, Liang, and Duchi]{carmon2022unlabeled}
Yair Carmon, Aditi Raghunathan, Ludwig Schmidt, Percy Liang, and John~C. Duchi.
\newblock Unlabeled data improves adversarial robustness, 2022.

\bibitem[Croce and Hein(2020{\natexlab{a}})]{croce2020minimally}
Francesco Croce and Matthias Hein.
\newblock Minimally distorted adversarial examples with a fast adaptive boundary attack.
\newblock In \emph{International Conference on Machine Learning}, pages 2196--2205. PMLR, 2020{\natexlab{a}}.

\bibitem[Croce and Hein(2020{\natexlab{b}})]{croce2020reliable}
Francesco Croce and Matthias Hein.
\newblock Reliable evaluation of adversarial robustness with an ensemble of diverse parameter-free attacks, 2020{\natexlab{b}}.

\bibitem[Dosovitskiy(2020)]{dosovitskiy2020image}
Alexey Dosovitskiy.
\newblock An image is worth 16x16 words: Transformers for image recognition at scale.
\newblock \emph{arXiv preprint arXiv:2010.11929}, 2020.

\bibitem[d’Ascoli et~al.(2021)d’Ascoli, Touvron, Leavitt, Morcos, Biroli, and Sagun]{d2021convit}
St{\'e}phane d’Ascoli, Hugo Touvron, Matthew~L Leavitt, Ari~S Morcos, Giulio Biroli, and Levent Sagun.
\newblock Convit: Improving vision transformers with soft convolutional inductive biases.
\newblock In \emph{International conference on machine learning}, pages 2286--2296. PMLR, 2021.

\bibitem[Foret et~al.(2021)Foret, Kleiner, Mobahi, and Neyshabur]{sam}
Pierre Foret, Ariel Kleiner, Hossein Mobahi, and Behnam Neyshabur.
\newblock Sharpness-aware minimization for efficiently improving generalization, 2021.

\bibitem[Goodfellow et~al.(2015)Goodfellow, Shlens, and Szegedy]{goodfellow2015explaining}
Ian~J. Goodfellow, Jonathon Shlens, and Christian Szegedy.
\newblock Explaining and harnessing adversarial examples, 2015.

\bibitem[Gopal et~al.(2024)Gopal, Yang, Zhang, Horton, and Chen]{clat}
Bhavna Gopal, Huanrui Yang, Jingyang Zhang, Mark Horton, and Yiran Chen.
\newblock Criticality leveraged adversarial training (clat) for boosted performance via parameter efficiency, 2024.

\bibitem[Han et~al.(2024)Han, Gao, Liu, Zhang, and Zhang]{peftlargemodels}
Zeyu Han, Chao Gao, Jinyang Liu, Jeff Zhang, and Sai~Qian Zhang.
\newblock Parameter-efficient fine-tuning for large models: A comprehensive survey, 2024.

\bibitem[He et~al.(2016)He, Zhang, Ren, and Sun]{he2016deep}
Kaiming He, Xiangyu Zhang, Shaoqing Ren, and Jian Sun.
\newblock Deep residual learning for image recognition.
\newblock In \emph{Proceedings of the IEEE conference on computer vision and pattern recognition}, pages 770--778, 2016.

\bibitem[Howard()]{imagenette}
Jeremy Howard.
\newblock Imagewang.

\bibitem[Hu et~al.(2021)Hu, Shen, Wallis, Allen-Zhu, Li, Wang, Wang, and Chen]{hu2021loralowrankadaptationlarge}
Edward~J. Hu, Yelong Shen, Phillip Wallis, Zeyuan Allen-Zhu, Yuanzhi Li, Shean Wang, Lu Wang, and Weizhu Chen.
\newblock Lora: Low-rank adaptation of large language models, 2021.

\bibitem[Krizhevsky and Hinton(2009)]{data}
Alex Krizhevsky and Geoffrey Hinton.
\newblock Learning multiple layers of features from tiny images.
\newblock 2009.

\bibitem[Li et~al.(2022)Li, Wang, Li, Xie, Sima, Lu, Qiao, and Dai]{li2022bevformer}
Zhiqi Li, Wenhai Wang, Hongyang Li, Enze Xie, Chonghao Sima, Tong Lu, Yu Qiao, and Jifeng Dai.
\newblock Bevformer: Learning bird’s-eye-view representation from multi-camera images via spatiotemporal transformers.
\newblock In \emph{European conference on computer vision}, pages 1--18. Springer, 2022.

\bibitem[Liu et~al.(2024)Liu, Wang, Yin, Molchanov, Wang, Cheng, and Chen]{dora}
Shih-Yang Liu, Chien-Yi Wang, Hongxu Yin, Pavlo Molchanov, Yu-Chiang~Frank Wang, Kwang-Ting Cheng, and Min-Hung Chen.
\newblock Dora: Weight-decomposed low-rank adaptation.
\newblock \emph{arXiv preprint arXiv:2402.09353}, 2024.

\bibitem[Liu et~al.(2021)Liu, Lin, Cao, Hu, Wei, Zhang, Lin, and Guo]{liu2021swin}
Ze Liu, Yutong Lin, Yue Cao, Han Hu, Yixuan Wei, Zheng Zhang, Stephen Lin, and Baining Guo.
\newblock Swin transformer: Hierarchical vision transformer using shifted windows.
\newblock In \emph{Proceedings of the IEEE/CVF international conference on computer vision}, pages 10012--10022, 2021.

\bibitem[Madry et~al.(2019)Madry, Makelov, Schmidt, Tsipras, and Vladu]{madry2019deep}
Aleksander Madry, Aleksandar Makelov, Ludwig Schmidt, Dimitris Tsipras, and Adrian Vladu.
\newblock Towards deep learning models resistant to adversarial attacks, 2019.

\bibitem[Mo et~al.(2022{\natexlab{a}})Mo, Wu, Wang, Guo, and Wang]{mo2022adversarialtrainingmeetsvision}
Yichuan Mo, Dongxian Wu, Yifei Wang, Yiwen Guo, and Yisen Wang.
\newblock When adversarial training meets vision transformers: Recipes from training to architecture, 2022{\natexlab{a}}.

\bibitem[Mo et~al.(2022{\natexlab{b}})Mo, Wu, Wang, Guo, and Wang]{recipe_trans}
Yichuan Mo, Dongxian Wu, Yifei Wang, Yiwen Guo, and Yisen Wang.
\newblock When adversarial training meets vision transformers: Recipes from training to architecture.
\newblock \emph{Advances in Neural Information Processing Systems}, 35:\penalty0 18599--18611, 2022{\natexlab{b}}.

\bibitem[Peebles and Xie(2023)]{peebles2023scalable}
William Peebles and Saining Xie.
\newblock Scalable diffusion models with transformers.
\newblock In \emph{Proceedings of the IEEE/CVF International Conference on Computer Vision}, pages 4195--4205, 2023.

\bibitem[Rice et~al.(2020)Rice, Wong, and Kolter]{rice2020overfitting}
Leslie Rice, Eric Wong, and J.~Zico Kolter.
\newblock Overfitting in adversarially robust deep learning, 2020.

\bibitem[Shafahi et~al.(2019)Shafahi, Najibi, Ghiasi, Xu, Dickerson, Studer, Davis, Taylor, and Goldstein]{shafahi2019adversarial}
Ali Shafahi, Mahyar Najibi, Amin Ghiasi, Zheng Xu, John Dickerson, Christoph Studer, Larry~S. Davis, Gavin Taylor, and Tom Goldstein.
\newblock Adversarial training for free!, 2019.

\bibitem[Su et~al.(2019)Su, Vargas, and Sakurai]{su2019one}
Jiawei Su, Danilo~Vasconcellos Vargas, and Kouichi Sakurai.
\newblock One pixel attack for fooling deep neural networks.
\newblock \emph{IEEE Transactions on Evolutionary Computation}, 23\penalty0 (5):\penalty0 828--841, 2019.

\bibitem[Szegedy et~al.(2014)Szegedy, Zaremba, Sutskever, Bruna, Erhan, Goodfellow, and Fergus]{szegedy2014intriguing}
Christian Szegedy, Wojciech Zaremba, Ilya Sutskever, Joan Bruna, Dumitru Erhan, Ian Goodfellow, and Rob Fergus.
\newblock Intriguing properties of neural networks, 2014.

\bibitem[Tian et~al.(2022)Tian, Wu, Dai, Hu, and Jiang]{tian2022deeper}
Rui Tian, Zuxuan Wu, Qi Dai, Han Hu, and Yu-Gang Jiang.
\newblock Deeper insights into the robustness of vits towards common corruptions.
\newblock \emph{arXiv preprint arXiv:2204.12143}, 2022.

\bibitem[Touvron et~al.(2021)Touvron, Cord, Douze, Massa, Sablayrolles, and J{\'e}gou]{touvron2021training}
Hugo Touvron, Matthieu Cord, Matthijs Douze, Francisco Massa, Alexandre Sablayrolles, and Herv{\'e} J{\'e}gou.
\newblock Training data-efficient image transformers \& distillation through attention.
\newblock In \emph{International conference on machine learning}, pages 10347--10357. PMLR, 2021.

\bibitem[Vaswani et~al.(2023)Vaswani, Shazeer, Parmar, Uszkoreit, Jones, Gomez, Kaiser, and Polosukhin]{vaswani2023attentionneed}
Ashish Vaswani, Noam Shazeer, Niki Parmar, Jakob Uszkoreit, Llion Jones, Aidan~N. Gomez, Lukasz Kaiser, and Illia Polosukhin.
\newblock Attention is all you need, 2023.

\bibitem[Xiao et~al.(2018)Xiao, Zhu, Li, He, Liu, and Song]{xiao2018spatially}
Chaowei Xiao, Jun-Yan Zhu, Bo Li, Warren He, Mingyan Liu, and Dawn Song.
\newblock Spatially transformed adversarial examples.
\newblock \emph{arXiv preprint arXiv:1801.02612}, 2018.

\bibitem[Xie et~al.(2020)Xie, Tan, Gong, Wang, Yuille, and Le]{xie2020adversarial}
Cihang Xie, Mingxing Tan, Boqing Gong, Jiang Wang, Alan Yuille, and Quoc~V. Le.
\newblock Adversarial examples improve image recognition, 2020.

\bibitem[Yang et~al.(2020)Yang, Zhang, Dong, Inkawhich, Gardner, Touchet, Wilkes, Berry, and Li]{yang2020dverge}
Huanrui Yang, Jingyang Zhang, Hongliang Dong, Nathan Inkawhich, Andrew Gardner, Andrew Touchet, Wesley Wilkes, Heath Berry, and Hai Li.
\newblock Dverge: diversifying vulnerabilities for enhanced robust generation of ensembles.
\newblock \emph{Advances in Neural Information Processing Systems}, 33:\penalty0 5505--5515, 2020.

\bibitem[Zagoruyko and Komodakis(2017)]{zagoruyko2017wide}
Sergey Zagoruyko and Nikos Komodakis.
\newblock Wide residual networks, 2017.

\bibitem[Zhang et~al.(2019)Zhang, Yu, Jiao, Xing, Ghaoui, and Jordan]{trades}
Hongyang Zhang, Yaodong Yu, Jiantao Jiao, Eric~P. Xing, Laurent~El Ghaoui, and Michael~I. Jordan.
\newblock Theoretically principled trade-off between robustness and accuracy, 2019.

\bibitem[Zhang et~al.(2024{\natexlab{a}})Zhang, Qiang, Somayajula, and Xie]{autolora}
Ruiyi Zhang, Rushi Qiang, Sai~Ashish Somayajula, and Pengtao Xie.
\newblock Autolora: Automatically tuning matrix ranks in low-rank adaptation based on meta learning, 2024{\natexlab{a}}.

\bibitem[Zhang et~al.(2024{\natexlab{b}})Zhang, He, Zhu, Chen, Wang, and Wei]{sam_and_trans}
Yihao Zhang, Hangzhou He, Jingyu Zhu, Huanran Chen, Yifei Wang, and Zeming Wei.
\newblock On the duality between sharpness-aware minimization and adversarial training.
\newblock \emph{arXiv preprint arXiv:2402.15152}, 2024{\natexlab{b}}.

\bibitem[Zhu et~al.(2023)Zhu, Wang, Hu, Xie, and Yang]{rift}
Kaijie Zhu, Jindong Wang, Xixu Hu, Xing Xie, and Ge Yang.
\newblock Improving generalization of adversarial training via robust critical fine-tuning, 2023.

\end{thebibliography}
